# A SURVEY OF METHODS FOR HANDLING DISK DATA IMBALANCE

Shuangshuang Yuan[1], Peng Wu[1], Yuehui Chen[1(✉)] and Qiang Li[2]

[1]School of Information Science and Engineering, University of Jinan, Jinan, China
yuanshuang1024@163.com, ise_wup@ujn.edu.cn, yhchen@ujn.edu.cn
[2]State Key Laboratory of High-end Server & Storage Technolog, Jinan, China
li.qiangbj@ieisystem.com

## ABSTRACT

*Class imbalance exists in many classification problems, and since the data is designed for accuracy, imbalance in data classes can lead to classification challenges with a few classes having higher misclassification costs. The Backblaze dataset, a widely used dataset related to hard discs, has a small amount of failure data and a large amount of health data, which exhibits a serious class imbalance. This paper provides a comprehensive overview of research in the field of imbalanced data classification. The discussion is organized into three main aspects: data-level methods, algorithmic-level methods, and hybrid methods. For each type of method, we summarize and analyze the existing problems, algorithmic ideas, strengths, and weaknesses. Additionally, the challenges of unbalanced data classification are discussed, along with strategies to address them. It is convenient for researchers to choose the appropriate method according to their needs.*

## KEYWORDS

*Classification, Disk Failure Prediction, Imbalanced Dataset, Data Processing*

## 1. INTRODUCTION

In the era of big data, the total volume of data is explosively expanding. Human social activities are increasingly reliant on digitized information, which necessitates robust support from massive storage systems. With the growing importance of data, the security of storage systems has become a major concern. Among the fundamental storage devices that still dominate the field is the hard disk, and once a disk is damaged, it can lead to permanent data loss. Therefore, disk failure prediction has become crucial.

As one of the mainstays of current smart processing technologies, machine learning provides important technical support for the aforementioned problems. Classification algorithms have become a key technique in machine learning to build efficient classifiers and extract effective information in data. When it comes to disk failure prediction, the analysis, drawn from the widely used public dataset Backblaze, reveals a serious category imbalance in the SMART dataset that reflects the health of disks. If the faulty samples (positive samples) are misclassified as normal samples (negative samples), which leads to the failure of correctly identifying the faulty disks, this will bring irreparable losses. In addition, existing classification models are mainly based on the assumption that the proportion of positive and negative samples is close and the cost of misclassification is comparable. However, in many problems, the data tends to present an imbalance of classes, i.e., the proportion of positive and negative samples is out of proportion, and there is a risk that all samples will be judged to be in the majority class by using a common and highly efficient classifier when dealing with data that has a significant imbalance of proportions [1]. Assuming an imbalance ratio of more than 20:1 for a dataset, the classifier can achieve a classification accuracy rate of 90 percent or even higher, but the accuracy is not of reference significance. Since the goal of the classification model is mainly based on the overall classification accuracy, the model will be biased towards the majority class and ignore the minority class, resulting in a small number of classes that can be excavated, resulting in low classification accuracy for the minority class. Thus, how to classify various types of imbalanced datasets has become a major area of research.

The problem of classification of imbalanced data is characterized by two main aspects: the imbalanced nature of the data set itself and the limitations of traditional classification algorithms. In terms of the characteristics of the data set itself, the problem of imbalanced classification of disk data is mainly manifested by the small number of disk fault sample data, which makes it difficult for the classifier to fully learn the feature information of the fault samples through training due to the limited information contained in the few classes themselves, making the disk fault data difficult to identify and thus misclassify causing loss. In the case of traditional classifiers, traditional classifiers are usually trained to minimize empirical or structural risk, to minimize misclassification rate, or to maximize inter-class intervals. When the data distribution is imbalanced, samples belonging to minority classes in overlapping parts of the class are misclassified in large numbers, and the class interval is shifted to the side of the class with a sparser sample distribution. Thus the classification accuracy of the algorithm for the minority class cannot be guaranteed [2].

In this paper, based on the research work related to the imbalance problem of disk dataset, Figure 1 summarizes the corresponding representative algorithms from three aspects: data-level methods (imbalanced data sampling methods), algorithm-level methods (improved algorithms based on machine learning) and hybrid methods, and also analyze the data imbalance classification problem, propose related challenges, and develop future research directions in this area.

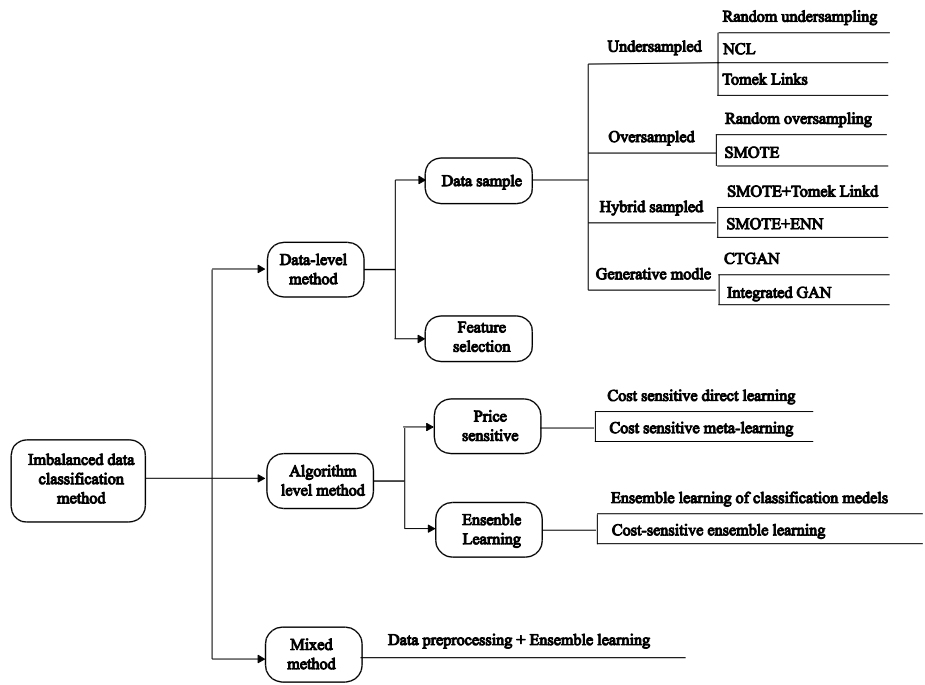

Figure 1. Imbalanced data approach

## 2. DATA-LEVEL METHOD

Most existing classifiers are based on the assumption that the positive and negative samples in the dataset are balanced, so a common approach to deal with data imbalance is to pre-process the imbalanced dataset so that it is balanced before training for classification. Data-level methods, also known as extrinsic methods, can be further subdivided into data sampling methods and feature selection methods, which are used to adapt the training set of a model at the level of data preprocessing, aiming to change the sample distribution of the training set by using resampling techniques [3-6] to reduce or eliminate the imbalance rate of positive and negative samples to make it suitable for the standard classification algorithm model. Data resampling methods: random undersampling, random oversampling and synthetic minority oversampling techniques are some commonly used algorithms in the MapReduce scheme in the framework of big data [7-9].

The methods to change the distribution of the dataset can be classified as generating minority class data samples (oversampling), removing majority class data samples (undersampling), mixed sampling and generating models. Oversampling methods and undersampling methods form the two sub-groups of the data-sampling methods group, where data sampling from the given dataset is done either randomly or using a specified formulaic/algorithmic approach [1,10]. Feature Selection Methods, while largely used only (without consideration to class imbalance) to improve classification performance [11,12], may also help select the most influential features (or attributes) that can yield unique knowledge for inter-class discrimination. This leads to a reduction of the adverse effects of class imbalance on classification performance [13-15].

## 2.1. Oversampling classification method

The simplest oversampling method is random oversampling, which balances the number of samples in both classes by replicating a few columns of samples, but the addition of a few classes of samples to enlarge the sample space leads to the generation of too much redundant and overlapping data, which leads to the overfitting of the classification model to the replicated samples. Most published works in the field of disk failure prediction that address the issue of data imbalance use data-level approaches. Kaur [16] oversampled a small number of samples. However, due to the high imbalance rate of positive and negative samples, a higher oversampling rate is adopted in this paper to increase the number of positive samples. The probability of recurrence of positive samples during oversampling is high, resulting in poor generalization performance. It can be seen that the improper use of oversampling deteriorates the prediction results and incurs increased computational cost. To address this issue, many scholars have improved the oversampling techniques and achieved results in real-world applications. Lin Yang [17] proposed to combine fuzzy classification with imbalanced data processing by introducing a bounded minority class to over samples near the decision boundary between majority class data and minority class data to generate minority class data, and a correlated fuzzy naive Bayes classifier was used to classify the data with satisfactory results by assigning a bounded balanced dataset to a probability index table. Luo Zhengbo [18] regarded the sample data at the classification boundary as error-prone data and oversampled them to obtain a new balanced dataset and used it to train the classifier, which improved the classification performance.

In addition to the above sampling improvements, Chawla [19] targeted a synthetic minority class sample-synthetic minorit oversampling technique (SMOTE), which analyzes real samples of minority classes and randomly selects samples of similar distance for interpolation to generate new samples of minority classes without duplicates, which can overcome the overfitting problem of random oversampling methods to a certain extent. SMOTE synthesizes sample data mainly using the K-nearest neighbor algorithm, which computes K-nearest neighbors for each minority class sample and randomly selects N samples from the K-nearest neighbors for random linear interpolation to construct a new minority class sample. The use of genetic algorithm (GA) to find the optimal sampling rate of SMOTE algorithm is a good research direction [20]. New samples created using SMOTE are sometimes less accurate because some minority classes tend to be present in regions where the majority class is distributed. Borderline SMOTE [21] is an improved oversampling algorithm based on SMOTE, which uses the same oversampling technique as SMOTE, but uses only a few classes of samples on the border to synthesize new samples, thus improving the class distribution of the samples. The sampling procedure of Borderline SMOTE is to split the minority sample into three classes, Safe, Danger and Noise, which are illustrated in Figure 2, and finally Borderline SMOTE oversamples the minority class sample of Danger only. Bunkhumpornpat [22] proposed the Safe-Level-SMOTE algorithm, which carefully samples minority class samples with different weights along the same line and assigns a security level to each original minority class sample before generating a minority class sample. It is experimentally demonstrate that this approach achieves better accuracy than SMOTE and Borderline-SMOTE.

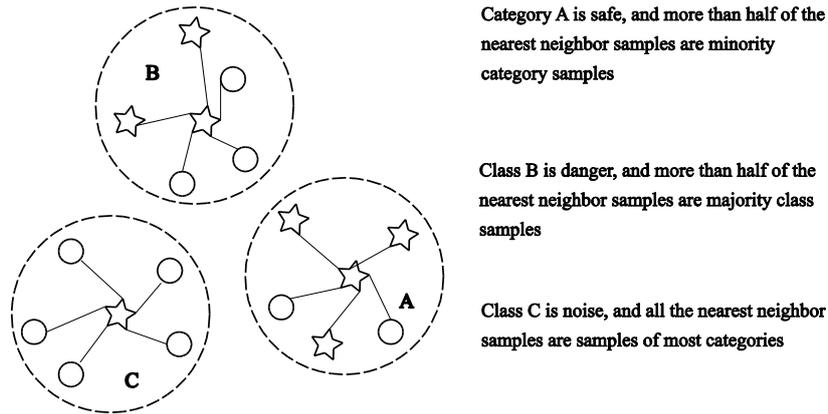

Figure 2. Borderline SMOTE:Classification of minority classess

## 2.2. Undersampling classification method

Undersampling, as a non-heuristic learning approach, can effectively alleviate the imbalance between majority and minority classes by randomly discarding some majority class data to build the classifier [23]. Li [24] heavily removed normal disk data (negative samples) to make the ratio of positive and negative samples 1:10 to alleviate positive and negative sample imbalance, random undersampling is the random removal of the majority class samples, where the majority class samples are often ignored and the classifier has difficulty in correctly learning the decision boundary between positive and negative samples due to the presence of data loss, which degrades the performance of the classifier. Therefore, data-level methods that use undersampling to balance the dataset need to focus on this problem and seek solutions from other techniques to compensate for this deficiency and improve the classification of imbalanced datasets. Heuristic undersampling methods have been proposed by related scholars, and the classical undersampling methods are Neighborhoodcleaning rule (NCL) and Tomek links method. Meng Dongxia [25] first used the K-nearest neighbor algorithm to remove the noisy points in a certain region and screen out the high-quality sample points, and then used the undersampling method to process the bounding-box majority samples and retain the high-quality samples in the process to reduce the sample information loss, but this method is prone to inaccurate noise discrimination in the process of screening noise. Regarding this kind of problem, the literature [26] used neighborhood density for data evaluation of the border majority class samples and used the majority class samples which are quantitatively equivalent to the minority class samples for data classification, effectively circumventing this problem.

## 2.2. Mixed sampling method

Both oversampling and undersampling methods have different drawbacks, and the combination of oversampling and undersampling methods provides a new direction for the imbalance classification problem in order to obtain better classification results. In the literature [27], an improved hybrid sampling strategy is proposed based on the nearest neighbors of the misclassified sample point data: the noisy samples are directly deleted; The dangerous samples are approximately removed from the majority class samples in their nearest neighbors and the safe samples are synthesized into new samples using the SMOTE algorithm and added to a new training set for re-training and learning. In [28], the SMOTE algorithm was improved by writing the Gini gain into the distance formula to alleviate the problem of oversampling to generate noisy data, and the samples are pre-processed using feature-weighted oversampling and data cleaning techniques, and we experimentally demonstrate that the recognition efficiency of the proposed algorithm improves after refinement and processing.

### 2.3. Generate model classification method

By reading the recent new publications, we found that in dealing with disk data imbalance Jia [29] used conditional table GAN (CTGAN) to generate data to expand the disk fault data, the table GAN generates synthetic data with similar distribution and the same properties as the real samples by learning from the real samples, and uses these generated synthetic data to balance the disk dataset, and after experimental comparison on some classification models, the experimental results show that CTGAN achieves better results by generating faulty samples to balance the disk data. Yuan [30] draw on the idea of ensemble learning to balance the disk dataset by using the failure samples generated by the ensemble mixed multivariate table GAN, and the experimental results show that the use of such mixed sample data can achieve more accurate disk failure prediction accuracy. The specific implementation flow of the disk failure prediction experiment using multivariate GAN to process imbalanced data in the literature [30] is shown in Figure 3. This hybrid approach not only increases the diversity of samples, but also smooths transitions between decision boundaries of different classes, reduces misclassification of few class samples, and makes the model more stable during training. Experimental results show that more accurate disk fault prediction accuracy can be achieved with such mixed-sample data.

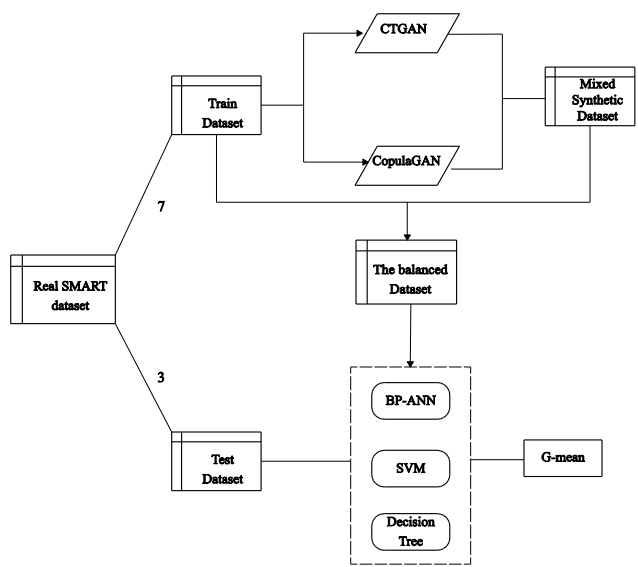

Figure 3. Experimental flow of Yuan [30] integrating GAN balanced data

Pre-processing of imbalanced disk datasets using data-level methods is independent of the specific classifier used and well-adapted. However, it is not very applicable when the imbalance rate of the data set is high, and choosing the optimal sampling rate when using oversampling methods for imbalanced data is a difficult task, and it is important to consider how to maintain the same spatial distribution of the data while reducing the imbalance ratio between positive and negative data samples when dealing with imbalanced data, Further combining the data distribution information, it is worth paying attention to determine the sampling rate of the best classification performance of resampling reasonably and effectively, and improve the classification performance of imbalanced data.

## 3. ALGORITHM-LEVEL METHOD

The problem of designing classification models that can overcome the negative effects of data imbalance has been widely discussed over the past two decades. We need to build classifiers that favor a few classes and do not affect the accuracy of the majority class. The algorithm-level approach is to make changes to the traditional classifier model to reduce the inherent classification bias of the classifier, rather than to the data distribution. The algorithm-level methods commonly used are the cost-sensitive method [31] and the integration method. Classifiers using cost-sensitive methods assign a different weight penalty to each input training sample. In this way, higher weights can be assigned to samples from minority classes, which increases the importance of minority classes during model training, biases the classifier towards minority classes, and reduces misclassification due to class imbalance.

Pereira [32] used classifiers such as Gaussian mixture model (GMM), K-nearest neighbor (KNN), SVM, and principal component analysis (PC) to predict disk faults. Jiang et al. Based on the idea of hybrid ensemble learning, Zhang [33] proposed a novel fault prediction method combining machine learning algorithms and neural networks on the publicly available BackBlaze hard drive dataset, obtaining an ensemble learning model with high performance on most types of hard drives. [34] used a long short-term memory neural network and a fully connected layer to design a self-encoder that can perform quadratic encoding and train the classification model using only normal samples, avoiding the negative effects due to the imbalance of positive and negative samples allows the model to learn the data distribution of normal samples, which improves the generalization ability of the model. Zhang [35] used an improved random forest-based disk fault prediction method, using the idea of integrated learning (Bagging algorithm) to construct multiple decision trees on the training set, and combined the classification voting results of multiple decision trees in the final prediction. It is proved that the random forest algorithm has good prediction accuracy, is robust to noisy data and outliers, and can avoid the overfitting phenomenon [36]. [37] and [38] integrate migration learning into classification models for disk failure prediction work with small samples of faulty disk data. Due to the rare occurrence of disk faults and the extreme imbalance between positive and negative samples, classifier models trained are likely to be overfitted. In the literature [37], among all data in the same time period, the disk data of several models with more faulty samples were selected from all disk models as the original domain training model, and by calculating the MMD distance between the original domain dataset and the target domain dataset, the original domain trained model with the smallest MMD is selected to continue training and testing with the target domain data, and it was found that the trained model using the original domain with more faulty samples was migrated to the target domain (with fewer faulty samples) and then classified the samples on the target domain much better than training the model directly on the target domain and then testing it. [38] used a threshold to define a minority class of disk data, i.e., a data set with less than 1500 faulty samples was defined as a minority disk instead of considering the ratio of positive and negative samples, and experiments showed that this model had an FDR of 96% and an FAR of 0.5%, and it was suggested in the paper that overfitting was associated with a small number of training samples, but overfitting was also affected by model complexity and improper data resampling [39].

The imbalanced data classification method based on cost-sensitive learning guides the learning process by introducing the cost, but when the number of minority samples is too small or even there are no minority samples, the cost-sensitive method is no longer applicable. Meanwhile, the core of the cost-sensitive learning based imbalanced data classification method is the determination of the misclassification cost, and it is difficult to make an accurate estimation of the true misclassification cost in most cases. A feasible solution is to take the best classification performance of the majority class and the minority class as the criterion, combine the distribution characteristics of the data to construct the cost distribution space, and use search and optimisation techniques to find a misclassification cost that best fits the characteristics of the dataset.

## 4. HYBRID-LEVEL METHOD

Hybrid methods are a combination of data-level and algorithm-level methods, drawing on the advantages and discarding the disadvantages [40]. Due to the continuous improvement of data-level methods and algorithm-level methods, hybrid methods are widely used, where the use of hybrid methods combines dataset methods with classifier integration [41] to produce classifiers with better classification results.

In many studies on disk failure prediction, hybrid methods have been used to achieve high classification accuracy and good classification results. The data-level approach used by Yunhua Song [42] in dealing with the disk data imbalance problem starts from two aspects: in the model training phase, the DKSS hybrid strategy (DBSCAN, K-means, SMOTE, and SVM) reconstructs the dataset by clustering and sampling methods to achieve a certain degree of balancing the ratio of positive and negative samples; In the model testing phase, the DKSS The hybrid strategy uses the clustering method to reduce the size of the samples to be tested in order to improve the detection rate, and the integration method is used in the algorithm level method to integrate the SVM classifier to predict the data samples, and the final classification prediction results are obtained by the voting method. A flowchart of the specific implementation is shown in Figure 4. Wang [43] proposed an adaptive data density-based oversampling technique (ASMOBD), which compared to existing resampling algorithms, can adaptively synthesize a different number of new samples around each minority sample according to the learning difficulty. Thus, this method makes the decision region more specific and can eliminate noise [43]. The combination of ASMOBD and the cost-sensitive SVM model with the cost-sensitive method applied in this paper achieves good results. Li [44] proposed an adaptive weighted Bagging-GBDT algorithm-based disk failure prediction model, where multiple sampling of normal samples is performed at the data level using hierarchical undersampling based on clustering; Multiple GBDT (gradient boosting decision tree) sub-classification models with high prediction accuracy are built at the algorithm level by training on a subset, and then the weights of each submodel are determined adaptively. Finally, the final disk failure prediction model is integrated by weighted hard voting. Tian [45] solved the sample imbalance problem by indirectly decreasing the weight of negative samples, and used an online disk failure prediction model based on LightGBM, which is an improvement of GBDT and solves the problem of GBDT consuming large computational and time costs, and achieves higher prediction accuracy while guaranteeing a low false alarm rate. Liu [46] used an oversampling method to balance the data set to increase a few classes; used the threshold method, Lasso feature selection to reduce the dimensionality of the data, and used the filtered data as the training set for training the XGBoost model, and compared the experiments with the rest of the classical machine learning algorithms (which include the LightGBM model), the experimental results of this paper show that LightGBM is overfitted when the sample is increased, and the test results are degraded. Combined with [45], it can be seen that disk failure prediction is a complex problem affected by multiple factors, and to design an accurate method for predicting disk failures, we need to consider various aspects such as feature selection, data preprocessing, model selection and improvement.

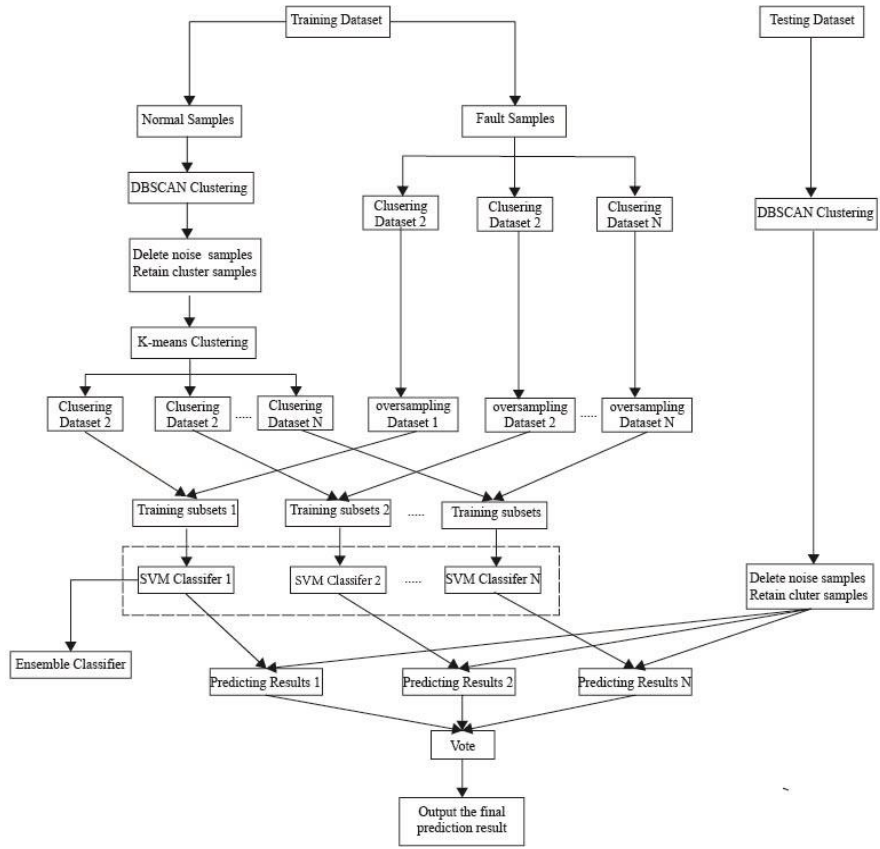

Figure 4. Flowchart of DKSS implementation by Yunhua Song [42]

## 5. CONCLUSION AND OUTLOOK

Data category imbalance is widespread, the disk data show obvious positive and negative sample category imbalance, due to the disk failure, bring more losses, so the treatment of imbalanced data has important research significance and application value. After a long period of development, the problem of dealing with imbalanced datasets has received a lot of attention, and a wide range of results have been achieved in various research areas. In this paper, we analyze data-level, algorithm-level, and hybrid approaches for solving the positive and negative sample imbalance problem in disk failure prediction and provide insights on future research priorities and directions.It provides researchers in the field of disk failure prediction with a reference method to analyse and help them to choose the appropriate method to deal with the problem of disk data imbalance according to their own needs.

1. Data category imbalance is widespread, the disk data show obvious positive and negative sample category imbalance, due to the disk failure, bring more losses, so the treatment of imbalanced data has important research significance and application value. After a long period of development, the problem of dealing with imbalanced datasets has received a lot of attention, and a wide range of results have been achieved in various research areas. In this paper, we analyze data-level, algorithm-level, and hybrid approaches for solving the positive and negative sample imbalance problem in disk failure prediction and provide insights on future research priorities and directions.It provides researchers in the field of disk failure prediction with a reference method to analyse and help them to choose the appropriate method to deal with the problem of disk data imbalance according to their own needs.

2. Algorithm-level approaches to deal with disk data imbalance: due to the nature of disk data itself, some disk types have high failure rates while others have few failures, but disk failure prediction is important for all disks. Therefore, using transfer learning to address the problem of poor classification prediction due to the imbalance of positive and negative samples in disk data is a good research direction; Alternatively, evolutionary learning and integrated machine learning models for predicting disk failure are also good research directions.

## ACKNOWLEDGEMENT

The research work in this paper was supported by the Shandong Provincial Natural Science Foundation of China (Grant No. ZR2019LZH003), Science and Technology Plan Project of University of Jinan (No. XKY2078) and Teaching Research Project of University of Jinan (No. J2158). Yuehui Chen is the author to whom all correspondence should be addressed.

**Authors**

**Shuangshuang Yuan** Currently, she is studying for a master's degree in Computer science from University of Jinan. Her main research interests are intelligent storage, data processing, and the application of deep learning.

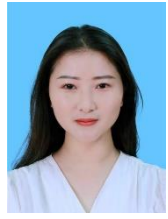

**Peng Wu** received Master from University of Jinan and a doctoral degree from Beijing Normal University. Currently, His main research interests are pattern recognition, bioinformatics, intelligent computing theory and application research.

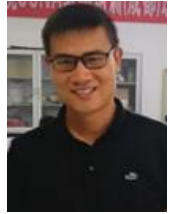

**Yuehui Chen** received his Ph.D. in Computer and Electronic Engineering from Kumamoto University, Japan. currently, His main research interests are theory and application of intelligent computing, bioinformatics and systems biology.

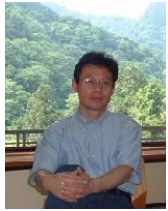

**Qiang Li** Currently, he is engaged in research on intelligent storage at the State Key Laboratory of High-End Server and Storage Technology.

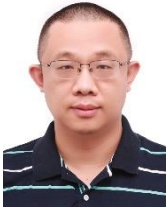